\documentclass[runningheads]{llncs}

\usepackage{makeidx} 
\usepackage{amssymb}
\usepackage{graphicx}
\usepackage{multirow}
\usepackage{hhline}
\usepackage{amsmath}
\usepackage{algorithmicx}
\usepackage{algorithm}
\usepackage{algpseudocode}
\usepackage{graphicx}

\begin{document}
	
\title{Privacy Information Classification: A Hybrid Approach}
\titlerunning{Privacy Information Classification: A Hybrid Approach}
%

\authorrunning{J.Q., W. et al.}

\author{Jiaqi Wu\inst{1} \and
Weihua Li\inst{1} \and
Quan Bai\inst{2} \and
Takayuki Ito\inst{3} \and
Ahmed Moustafa\inst{3}}

\institute{	
	Auckland University of Technology, Auckland, New Zealand, \\
	\and University of Tasmania, Tasmania, Australia \\
	\and Nagoya Institute of Technology, Japan \\
	\email {zcf3888@autuni.ac.nz, weihua.li@aut.ac.nz,quan.bai@utas.edu.au, ito.takayuki@nitech.ac.jp, ahmed@nitech.ac.jp}
}

%

%

\maketitle              
\begin{abstract}

A large amount of information has been published to online social networks every day. Individual privacy-related information is also possibly disclosed unconsciously by the end users. Identifying privacy-related data and protecting the online social network users from privacy leakage turn out to be significant. Under such a motivation, this study aims to propose and develop a hybrid privacy classification approach to detect and classify privacy information from OSNs. The proposed hybrid approach employs both deep learning models and ontology-based models for privacy-related information extraction. Extensive experiments are conducted to validate the proposed hybrid approach, and the empirical results demonstrate its superiority in assisting online social network users against privacy leakage. 


\end{abstract}
\section{Introduction}

With the proliferation and popularisation of the World Wide Web, Online Social Networks (OSNs) become of one of the essential channels for social interactions and communications \cite{batra2018characteristics,cormode2008key}. OSNs provide great convenience to the users, but these online social platforms also raise potential risks, such as privacy leakage. A vast amount of private information can be accessed publicly through OSNs, such as preferences, email address, marital status, hometown, activities attended, etc., which may lead to severe security issues.  


Therefore, it is significant to explore a useful and practical approach to protect OSN users from privacy information disclosure. Users should be reminded before posting any privacy-related messages to the public. In the context of OSNs, ``user privacy" refers to a sequence of words, stating or implying any individual' s personal information, preferences, events that he or she involved; privacy leakage describes a situation when an individual shares stories including private information with their contacts or even those they are not familiar with. Thus, it is necessary to develop a tool to detect and identify all the possible privacy-related information contained in any posting messages \cite{wang2011regretted,hasan2013discussion}. More importantly, the justifications of privacy-related information classification would be helpful for OSN users to get rid of posting similar messages again.  



In our previous research work \cite{li2019automated}, we conducted preliminary studies and developed a generic framework of privacy leakage detection for OSN users based on deep learning models. The proposed framework is capable of capturing privacy-related entities after giving sufficient training. However, two significant limitations are to be covered. Firstly, it can only remind users regarding the possibility of privacy leakage. As a result, detailed leaking information in terms of what kind of leakage is missing. Secondly, the privacy model is not conceptually modelled or presented.


Ontology models conceptually reflect the domain-specific knowledge in the form of terms and demonstrate two apparent advantages, i.e., shareability and reusability \cite{zhao2009ontology}. Therefore, ontology-based privacy models can be easily extended and applied to various OSNs \cite{mitra2005privacy}. Moreover, as ontology organises the concepts in the form of taxonomy or hierarchy based on a pre-defined natural relationship, it is suitable to introduce ontology into the research of privacy-related information extraction and classification.

As an extension of our previous work \cite{li2019automated}, in this paper, we leverage a hybrid privacy classification approach, incorporating both deep learning and ontology models, for individual users of OSNs. The extended framework is capable of addressing the privacy leakage problem for individuals by effectively identifying privacy information and classifying into a detailed category. More specifically, the proposed hybrid approach is composed of two major components, i.e., a deep learning based approach to detect the privacy leakage on online social data and an ontology privacy model classifying the privacy-related information into fine-grained privacy categories. Deep learning models are utilised to conduct the Name Entity Recognition (NER) and detect the pre-defined privacy-related entities. An ontology model is developed based on the analysis of massive data collected from real-world OSNs. Given the predictive results carried out by the deep learning model, the privacy ontology model further classifies the recognised privacy-related entities into sub-classes.

The rest of the paper is organised as follows. Section 2 reviews the existing literature of privacy information classification approaches on OSNs. Section 3 introduces the entire privacy information detection and fine-grained classification approach. In Section 4, experiments are conducted to evaluate the ontology-based approach on the dataset crawled from twitter. Section 5 concludes the findings of this paper and points out the limitations and future direction.

\section{Related Work}

Most contemporary privacy information classification approaches aim to detect some specific categories of privacy information on OSNs or perform a binary classification, i.e., sensitive or non-sensitive, rather than identifying and classifying privacy information for the end users. \cite{humphreys2010much} indicates that it is easy for users to leak the matters or activities that one involves anytime anywhere unconsciously. For example, a tweet saying that user’s family will go out for a holiday implies no one stays at home, which may cause robbery accidents. Therefore, the privacy information of OSNs users should be prompted before sharing with the public. 




\cite{gomez2010data} proposes a machine learning classification approach by adopting Named Entity Recognition technology, which can classify privacy information on OSNs into different categories, e.g., electronic devices and brands. Similarly, \cite{mao2011loose} present a privacy classifier for three kinds of sensitive tweets, i.e., drunk, holiday and disease, and classifies tweets into binary results, i.e., sensitive tweets or nonsensitive tweets.




From the research above, we can see even there are several papers about classifying privacy revealing information on OSNs, most of them pay attention to specific categories by machine learning approaches. Very few studies classify them according to a domain privacy ontology and protect individual OSNs users from online privacy leaks. Consequently, an ontology-based classification approach is comparatively a new research field of privacy information classification on OSNs. Ontology-based classification approaches show outstanding performance in many areas, e.g., online job offers \cite{ul2018ontology} and trust requirements on semantic web services \cite{galizia2006wsto}. Whereas, it is necessary to build an ontology into the OSNs \cite{kim2002building}.

\cite{gharib2017towards} survey the privacy ontology and point out that although some security ontologies for fulfilling security requirements have been presented, these studies focus on security rather than privacy \cite{souag2012ontologies}. They present a novel privacy ontology to identify the key concepts and relations to satisfy the privacy requirement. However, it aims to deal with the privacy requirements for software engineers. Actually, there is little research about privacy information ontology which can apply for OSNs.

Therefore, in this paper, we aim to build an ontology about privacy information on online social media. Because almost all the privacy-related entities are words or phrases, the semantic similarity degree calculating plays a significant role in the proposed ontology-based privacy approach.

\section{Automated Hybrid Privacy Detection Framework}

\begin{figure}
\includegraphics[width=\textwidth]{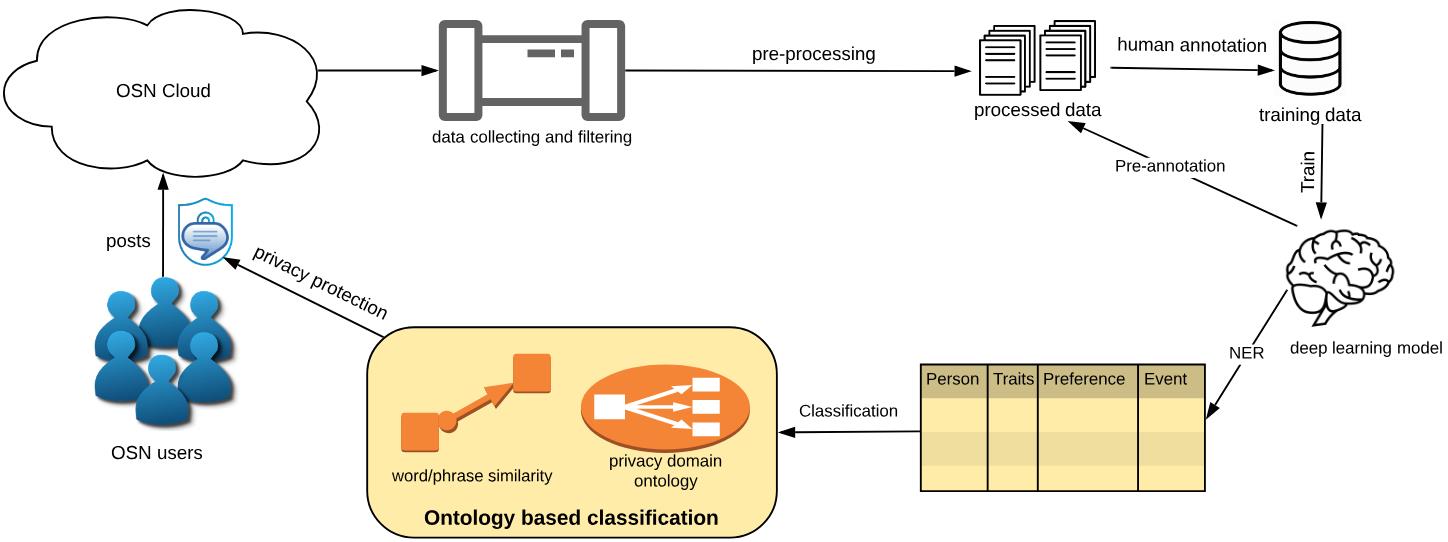}
\caption{Automated Hybrid Privacy Detection Framework}
\label{fig:framework}
\end{figure}

The proposed automated hybrid privacy information detection and classification approach are demonstrated in Fig \ref{fig:framework}. There are two key parts in the proposed framework, i.e., the privacy-related entities recognition and further classification based on ontology models. In the former, a deep learning model is trained to detect four types of entities that potentially cause privacy leakage. While, in the latter, a privacy ontology model is developed based on the analysis of messages posted by the OSN users.

The rationale of utilising a hybrid model of both deep learning and ontology-based classification is clarified as follows: ontology-only approach classifies privacy information merely based on the domain-specific vocabulary of terms or concepts, which have to be systematically defined \cite{noy2001ontology}. Deep learning based NER does not necessarily require a lexicon or domain-specific words, but it is difficult to control the classification of specific terms as deep learning models turn out to be a ``black-box". By considering the factors mentioned above, a combination of deep learning and ontology model is adopted.

The details of these two key modules are introduced in the following subsections.


\subsection{Deep-learning Based Privacy Information Detection}

Users' data collection turns out to be the preliminary of deep-learning based privacy information detection. OSN users' public data can be collected from OSNs through web crawlers or available APIs. The collected raw data are supposed to be filtered and pre-processed, e.g., removing messages which are advertisements or spams, removing meaningless words and characters and parsing word sequences to tokens. 

Next, the processed data are enriched by running through the pre-annotation process if a privacy-detection NER model is already available. Given the pre-annotated dataset, privacy-related entities are required to be annotated manually, and the data size of the training and testing data for the deep learning model are nearly 20k. As mentioned previously, in the context of OSNs, ``privacy" is associated with an individual' s personal information, preferences, events that he or she involved. Thus, messages containing four types of entities, i.e., ``PERSON", ``TRAIT", ``PREF", and ``EVENT", potentially cause privacy leakage. After annotation, the annotated dataset is then fed into the deep learning model for training.

\subsection{Ontology-based Privacy Information Classification}

As the second part of our hybrid privacy detection approach, an ontology-based classification approach can remind individual users regarding what is to be disclosed instead of simply giving general information. 

\subsubsection{The Domain and Scope of The Privacy Domain Ontology}

The defining domain and scope of ``privacy” is the first step to build a private-related information ontology model \cite{noy2001ontology}. Naturally, the privacy information concepts describing different subclasses (class corresponds to the entity in this thesis) of the four privacy-related entities will be fed into our ontology. Specifically, the privacy domain ontology on OSNs includes:

\begin{enumerate}
    \item A hierarchical classification of privacy concepts from general classes to specific subclasses.
    \item A set of relations between privacy classes to link concepts in a more complicated way that implied by an underlying hierarchy.
\end{enumerate}

\subsubsection{Privacy-related Keywords Extraction}

Initially, it is significant to obtain a comprehensive list of privacy-related terms and concepts in order to form the hierarchy of privacy ontology. To construct an ontology, we extracted all the values of privacy-related entities recognized by the deep learning based model. Next, based on the word frequencies, representative keywords are selected as the major indicators of the subcategories of the entities. For example, we want to find some keywords regarding private events under the ``Event" main class. Some verbs representative of private events, e.g., eating, shopping, etc., as well as some nouns, e.g., concert, meeting, journey, etc., are frequently mentioned in event-related entities. Additionally, some words are significant indicators of privacy-related entities, which can imply the user is leaking his/her ’TRAIT’ sensitive information.

Because the creation process of an ontology is an interactive process \cite{madsen2006health}, we searched the selected keywords in the dataset to find out the occurrence of these words and how important of them according to the term frequency \cite{noy2001ontology}. Through the interactive procedure, the terms and concepts of the privacy information ontology are finally determined. For example, because the keyword of ``interview" frequently appears in the extracted entities, it turns out to be a keyword, representing the subclass of ``Corporate Event".

Table~\ref{tab1} shows the representative keywords extracted from the collected data, which are used for building the privacy domain ontology. 

\begin{table}
\caption{Corresponding Keywords with Classes and Subclasses}\label{tab1}
\begin{tabular}{ | l | l | p{8cm} |}  
\hline
{\itshape Class}    & {\itshape Subclass} & {\itshape Keywords} \\
\hline
Person      & Individual    & I      \\
            & Third Party   & you, we, they, he, she, classmate, uncle \\
\hline
Preference  & Item          & book, chocolate, keyboard, tea     \\
            & Hobby         & cosplay, paint, fishing, dancing, reading  \\
            & Specific Person    & girlfriend, teacher  \\
\hline            
Event       & Private Event    & eat, shopping, concert, movie, exercise, spa  \\
            & Corporate Event  & wedding, interview, meeting, conference, festival, party, parade, salon\\
            & Journey          & fly, holiday, travel, island, hotel, airport\\
\hline            
Trait      & Individual Identity    & years-old, Auckland \\
            & Linked Information   & lawyer, female, gay, Christian, married, white, disable \\
\hline
\end{tabular}
\end{table}

\subsubsection{Privacy Ontology}

Among the possible approaches in developing a class hierarchy, a top-down process has been selected by considering the relationships among the privacy concepts in this paper \cite{noy2001ontology}. The ontology hierarchy presents a tree structure, having most general classes on top and specific associative classes connected with the general ones. For example, given ``PERSON" as a superclass, ``Individual" and ``Third Party" can be the subclasses based on the recognised entity-value pairs. Three other superclasses, i.e., ``TRAIT", ``EVENT", and ``PREFERENCE", are also included.

TRAIT describes personally identifiable information(PII) or sensitive personal information, which is defined and classified into two types from the usage of the PII in United Stated legal fields, i.e., distinguish identity and relating information \cite{mccallister2010sp}. Similarly, TRAITS can be classified into two subclasses as below:

\begin{enumerate}
    \item Any information can be used to distinguish an individuals identity, such as birth date and hometown.
    \item Any information can be linked to an individual, such as medical, educational, marital status and employment information.
\end{enumerate}

Through this kind of classification approach, two subclasses of Trait are identified: Individual identity and Linked information. For example, the date of birth can be recognised as an individuals identity. Whereas, race, gender, sexual orientation, marital status, religion, belief, and education background are categorized as linked information. Similarly, EVENT can be classified as a private event, corporate event, and journey in terms of the event is social or non-social. Among the subclasses of ``Event", tweets about the journey is individually classified because we think users who reveal their journey plans will make them very vulnerable to theft crimes. Preference can be classified as a specific person, item and hobby according to the characteristic of the leaking hobby information. Therefore, the privacy domain ontology for OSNs is presented in Fig.~\ref{fig:Ontology}:

\begin{figure}[ht] 
   \centering
   \includegraphics[width=1.2\linewidth]{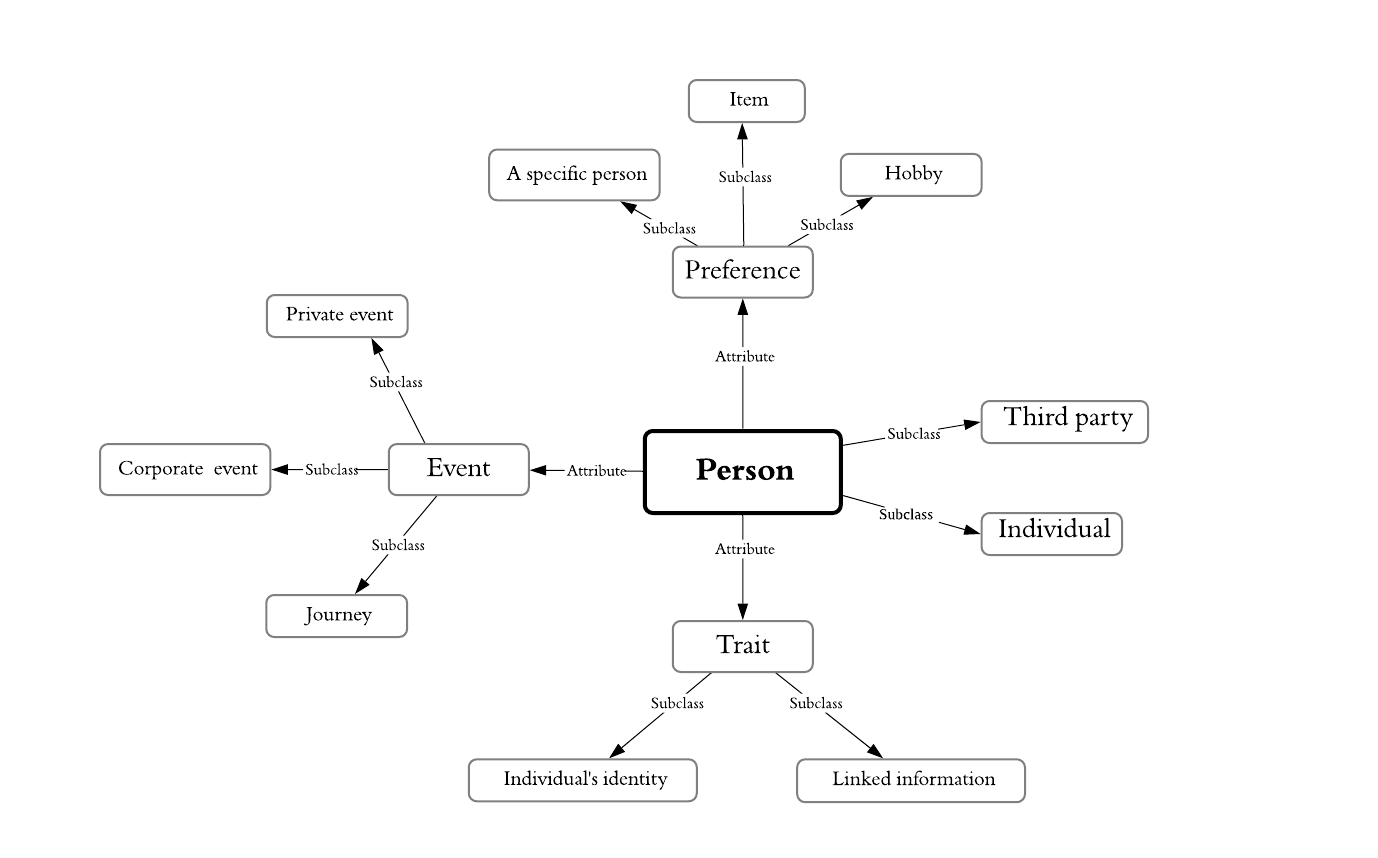} 
   \caption[Privacy Ontology]{Privacy Ontology}
   \label{fig:Ontology}
\end{figure}

\subsubsection{Semantic Phrase Similarity Degree Based On GloVe}

To classify the privacy-related entities into fine-grained subclasses, it is essential to find an approach in finding the subclass where the entities have the highest belonging degree. Moreover, the highest belonging degree is decided by the highest similarity degree between the extracted entities and the representative terms. Comparing the similarity of word embedding of each word can decide their similarity degree \cite{jiang1997semantic}. Word embedding is defined as semantic vector space models that use vectors to represent each word. 

Consequently, in this subsection, we will propose a semantic phrase similarity degree approach based on Glove. It is divided into two steps. Firstly, a word semantics vector space model is decided. Secondly, the construction of the classification model will be described by taking ``Event" entity as an example.

$\bullet$ Word Semantics Vector Space Model

In the first step, the main step is to choose a word semantics vector space model. A pre-trained statistical model (called ``en\_core\_web\_lg" in spacy) is used in this paper, which is trained on blogs, news, and comments with GloVe. Global Vectors for Word Representation (GloVe) is a state-of-art tool using word embedding techniques. Most word embedding approaches like Word2Vec exist a disadvantage, which is the lack of co-occurrence between words. Luckily, the GloVe approach trains global word-word co-occurrence counts which fills this gap, outperforming than other current word embedding approaches in common word similarity tasks\cite{pennington2014glove} \cite{mihalcea2006corpus}. That is the reason we choose this word semantics vector space model for us to use in order to compare the similarity degree between words.

$\bullet$ The Construction of The Classification Model

To explain the construction of the classification model, we make use of ``Event" entities as an example. In Fig.~\ref{fig:Ontology}, the entity consists of three subclasses, ``Private Event", ``Corporate Event" and ``Journey". Moreover, in Table~\ref{tab1}, there are 20 representative terms representing the subclasses associated with them. Each representative term can be represented as \(term_{i}\), where \(i\) belongs to \{0,1,2,...,19\}. Among them, ’Private Event’:\{0,1,2,...,5\}, ’Corporate Event’:\{6,7...,13\}, and ’Journey’:\{14,15...,19\}. Similarly, each word in the extracted entities can be represented as \(entity_{j}\).


Consequently, \(S_{ij}\) can represent the semantic similarity degree between each word in the extracted entities and each representative term:

\begin{equation}
S_{ij}= similarity(entity_{j},term_{i})
\end{equation}

So the similarity of event-related entities and the event representative terms in the ontology can be shown as follows:

\begin{equation}
S_{i(sum)}=\sum_{j=1}^{n}S_{ij}
\end{equation}

After the calculation of all the similarity degree, then the subclass of the extracted event entities can be decided according to the maximum degree of \(S_{i(sum)}\), which means the subclass which obtains the maximum \(S_{i(sum)}\) is the corresponding subclass of the event-related entity:

\begin{equation}
\lambda=\max \left(S_0{(sum)}, S_1{(sum)},...,S_{19}{(sum)}\right)  
\end{equation}

Then the \(i\) which obtains the maximum degree of \(S_{i(sum)}\) can be decided and the corresponding subclass can be decided as below:

$$
subclass = \left\{ \begin{array}{ll}
Private Event & \textrm{$i\in {0,1,...,5}$}\\
Corporate Event & \textrm{$i\in {6,7,...,13}$}\\
Journey & \textrm{$i\in {14,15,...,19}$}
\end{array} \right.
$$


Hence, by using the semantic information which the word embedding technique capture \cite{yaguinuma2010model}, we can classify privacy-related information into the corresponding subclass in the privacy ontology like the ``Event" procedure we list.

\section{Experiments}

Experiments have been conducted to evaluate the proposed hybrid privacy detection approach. The experiment uses a real-world testing dataset that aims to classify fine-grained privacy-related entities. Moreover, we also focus on one interesting problem around private tweets: What type of personal private information is leaked most on OSNs?

\subsection{Data Description}

Twitter \footnote{https://twitter.com/} is one of the largest Online Social Media platforms as a micro-blogging service, where a large amount of information is broadcast publicly by individual users. In Twitter, the information posted by end users is named as tweets. Twitter provides APIs, allowing developers to search and store tweets based on certain criteria. Therefore, we search and collect 18k tweets from Twitter through API. Most tweets are selected by searching for keywords related to sensitive activities and plans. 

Consequently, we use Twitter API to search for some terms which contain sensitive keywords, e.g., sensitive activities and plans, which may result in privacy information leakage and cause negative consequences. Then we collect around 18k tweets as a testing dataset in this experiment. As we demonstrated before, the tweets will be conducted with the deep learning based NER approach and the ontology-based classification approach in this chapter, then the corresponding subclasses will be decided.

\subsection{Evaluation}

To evaluate the performance of the hybrid privacy detection approach, we manually annotate the subclasses of privacy leaking information. Moreover, the ”ground truth” is prepared by allowing the users themselves to provide opinions on whether they leak the privacy and what types of private information they are leaking on tweets. For example, a tweet ``I watch a movie.", has ``I" annotated as ``Individual" and ``watch a movie" annotated as ``Private Event". 

Four traditional measures to utilised to evaluate the performance of the proposed hybrid approach: 

\begin{enumerate}
    \item Accuracy: the fraction of correct classification in the testing data set;
    \item Precision: the fraction of correct classification among all results are classified in this subclass in the testing data set;
    \item Recall: the fraction of correct classification among all actual results belong to this subclass in the testing data set;
    \item F1-value: the harmonic value of precision and recall, which is a balance measurement.
\end{enumerate}



\subsection{Experimental Results}

Our hybrid privacy classification approach utilizes a deep learning based NER approach and an ontology-based approach to perform classification of specific privacy information. After the NER, we use the ontology vocabulary for performing semantic phrase similarity degree calculation with the extracted privacy-related entities. 

We evaluate our hybrid approach on the testing dataset and get a considerable performance, as shown in Table~\ref{tab2}, with high accuracy in each type. We believe this accuracy is high enough to demonstrate the effectiveness of our automated detection and classification approach. However, we observe the accuracy of categories under ``TRAIT" entity is much lower than other types of entities. The extraction result of ``TRAIT" entity is lower than other types of entity and a trait-related privacy entity will be classified to categories under other types of entities. That is why the accuracy value of classification of ``TRAIT" entity is lower than other entities.

\begin{table}
\caption{Performance of Hybrid Privacy Information Classification}\label{tab2}
\begin{tabular}{| p{4cm} | p{4cm} | p{4cm} |}  
\hline
{\itshape Class}    & {\itshape Subclass} & {\itshape Accuracy} \\
\hline
Person      & Individual    &  0.94    \\
            & Third Party   &  0.85    \\
\hline
Preference  & Item          &   0.82   \\
            & Hobby         &   0.81   \\
            & Specific Person    & 0.76   \\
\hline            
Event       & Private Event    &  0.74  \\
            & Corporate Event  &  0.76  \\
            & Journey          &  0.77  \\
\hline            
Trait      & Individual Identity    & 0.62 \\
            & Linked Information   &  0.64 \\
\hline
\end{tabular}
\end{table}

\subsection{Privacy Leaking Information Categorization}

\begin{figure}[ht] 
   \centering
   \includegraphics[width=1.0\linewidth]{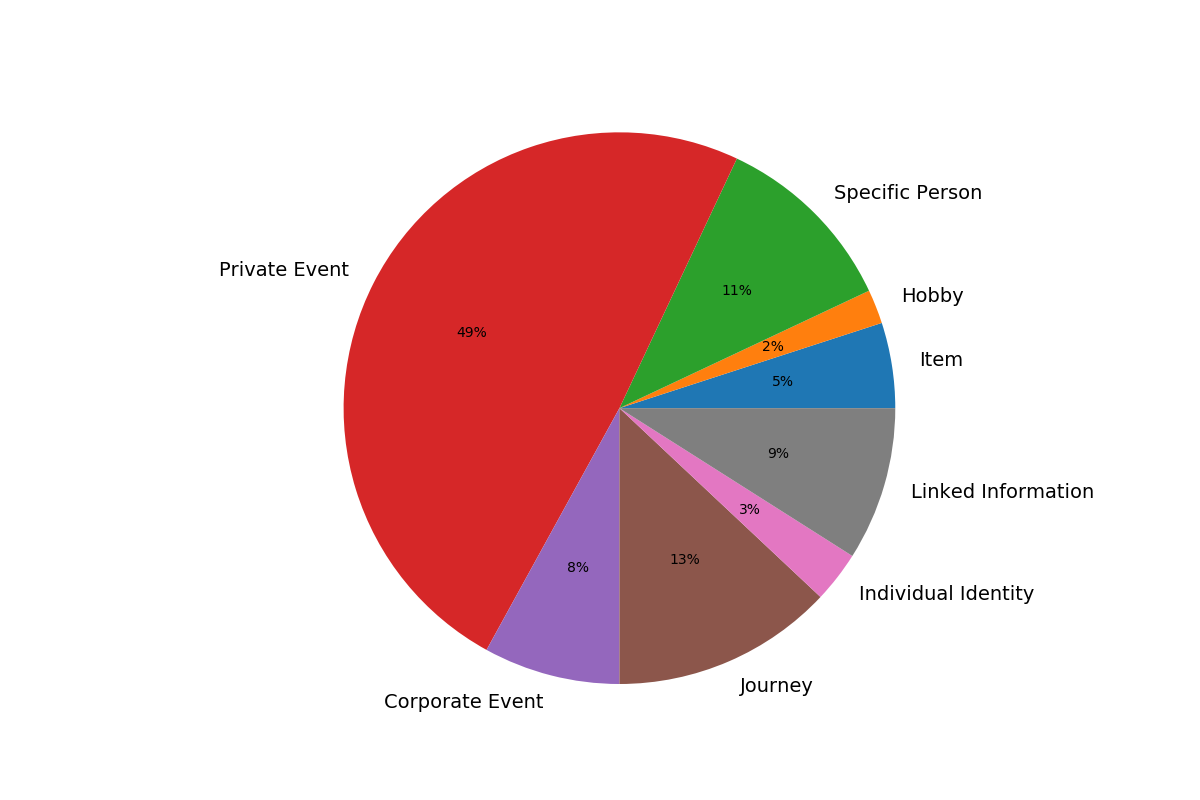} 
   \caption[Distribution of Different Types of Privacy Information Leaking]{Distribution of Different Types of Privacy Information Leaking}
   \label{fig:distribution}
\end{figure}

After the whole automated hybrid privacy information detection approach, all the types of private information leaks on the 18k testing dataset can be extracted. Then we plot the distribution of the results of the privacy information types of the testing dataset in Fig.~\ref{fig:distribution}, which includes all the subclasses under the ``TRAIT", ``PREF", and ``EVENT" entities. According to the privacy rule in this detection approach, each tweet with privacy-related entities which is identified as the private tweet contains the ``Person" entity, so the distribution of the ``Person" entity is not necessary to be analyzed its categorization.

In this section, we also explore the question of what type of sensitive information users leak most. We counted the percentage of the eight types of privacy information in the 18k testing dataset. The results are shown in Fig.~\ref{fig:distribution}. From Fig.~\ref{fig:distribution}, we can see the most leaking information is the ``Event" entity, where ``Private Event" counts the most in it. Additionally, ``Journey" under the ``Event" and ``Specific person" under the ``Preference" are also leaked a lot in tweets, which means the information about all subclasses under the entity of ``Event" both is leaked a lot on OSNs. From the results, we suggest that OSNs users should exercise a little more restraint about posting relevant tweets about these types. On the other hand, people are more conservative about ``Individual Identity" and ``Hobby" information because the privacy leaks (as the percentage) of the two types is much smaller than other types.


\section{Conclusions and Future Works}

In this paper, we propose a hybrid approach for classifying private information generated by OSN users. Through characterising the nature of privacy information leaking on OSNs, a deep learning model has been employed for privacy-related entities recognition, and an ontology-based classification approach is conducted to automatically classify fine-grained privacy information, i.e., nine subtypes of private leaking. The ontology-based approach calculated the semantic similarity between entities extracted from the deep learning model and the representative terms. We evaluated the result with the accuracy value, which demonstrates it gains a considerable performance. Moreover, what specific types of personal private information users are leaking on OSNs can be understood.


This research can be extended by investigating the following directions. Firstly, we can recognise privacy-related entities on tweets other than on all tweets by one user. In the future, all privacy information revealing by a user can be collected by the model and protect the user. Secondly, different tweets are associated with different degrees of privacy leakage. In this paper, we demonstrate what types of privacy-related entities OSNs users reveal. However, we do not analyse the detailed privacy leakage degree of users. In the future, we plan to evaluate the privacy leakage degree and explore the insights based on predictive results. 

\bibliographystyle{splncs04}
\bibliography{bibfile}
%





\end{document}